\documentclass[10pt,conference,a4paper]{IEEEtran}
\ifCLASSINFOpdf
\else
\fi
\usepackage{graphicx}
\usepackage{amsmath}

\usepackage[utf8]{inputenc}
\usepackage{xcolor}
\usepackage{subfig}
\usepackage[hidelinks]{hyperref}

\usepackage{booktabs}
\usepackage[shortcuts,acronym]{glossaries}

\usepackage{amssymb}
\usepackage{textcomp}

\usepackage{siunitx}
\newacronym{lasso}{LASSO}{least absolute shrinkage and selection operator}  
\newacronym{nwp}{NWP}{numerical weather prediction}
\newacronym{ml}{ML}{machine learning}
\newacronym{svr}{SVR}{support vector regression}
\newacronym{svm}{SVM}{support vector machine}
\newacronym{mlp}{MLP}{multi-layer perceptron}
\newacronym{gbrt}{GBRT}{gradient boosting regression tree}
\newacronym{mae}{MAE}{mean absolute error}
\newacronym{rmse}{RMSE}{root mean squared error}
\newacronym{mse}{MSE}{mean squared error}
\newacronym{kld}{KLD}{Kullback-Leibler Divergence}
\newacronym{pv}{PV}{photovoltaic}
\newacronym{wrt}{w.r.t.}{with respect to}
\newacronym{pdf}{PDF}{probability density function}
\newacronym{mtl}{MTL}{multi-task learning}
\newacronym{stl}{STL}{single-task learning}
\newacronym{sps}{SPS}{soft parameter sharing}
\newacronym{hps}{HPS}{hard parameter sharing}
\newacronym{csn}{CSN}{cross-stitch network}
\newacronym{sn}{SN}{sluice network}
\newacronym{ern}{ERN}{emerging relation network}
\newacronym{nlp}{NLP}{natural language processing}

\newcommand\SEC[1]{Sec.~\ref{#1}} 
\newcommand\FIG[1]{Fig.~\ref{#1}} 
\newcommand\TBL[1]{Table~\ref{#1}} 

\begin{document}

%
\title{Emerging Relation Network and Task Embedding for Multi-Task Regression Problems}

\author{\IEEEauthorblockN{Jens Schreiber}
\IEEEauthorblockA{University of Kassel\\
Willhelmshöher Allee 71-73\\
34121 Kassel\\
Email: j.schreiber@uni-kassel.de}
\and
\IEEEauthorblockN{Bernhard Sick}
\IEEEauthorblockA{University of Kassel\\
Willhelmshöher Allee 71-73\\
34121 Kassel\\
Email: bsick@uni-kassel.de}}


%


\maketitle

\begin{abstract}
  \Ac{mtl} provides state-of-the-art results in many applications of computer vision and natural language processing. 
  In contrast to \ac{stl}, \ac{mtl} allows for leveraging knowledge between related tasks improving prediction results on the main task (in contrast to an auxiliary task) or all tasks. 
  However, there is a limited number of comparative studies on applying \ac{mtl} architectures for regression and time series problems taking recent advances of \ac{mtl} into account. 
  An interesting, non-linear problem is the forecast of the expected power generation for renewable power plants. Therefore, this article provides a comparative study of the following recent and important \ac{mtl} architectures: 
  \Ac{hps}, \ac{csn}, \ac{sn}.
  They are compared to a \ac{mlp} model of similar size in an \ac{stl} setting.
  Additionally, we provide a simple, yet effective approach to model task specific information through an embedding layer in an \ac{mlp}, referred to as task embedding.
  Further, we introduce a new \ac{mtl} architecture named \ac{ern}, which can be considered as an extension of the \ac{sn}. 
  For a solar power dataset, the task embedding achieves the best mean improvement with $\mathbf{14.9\textbf{\%}}$. 
  The mean improvement of the \ac{ern} and the \ac{sn} on the solar dataset is of similar magnitude with $\mathbf{14.7\textbf{\%}}$ and $\mathbf{14.8\textbf{\%}}$. On a wind power dataset, only the \ac{ern} achieves a significant improvement of up to $\mathbf{7.7\textbf{\%}}$. Results suggest that the \ac{ern} is beneficial when tasks are only loosely related and the prediction problem is more non-linear. Contrary, the proposed task embedding is advantageous when tasks are strongly correlated. Further, the task embedding provides an effective approach with reduced computational effort compared to other \ac{mtl} architectures. 
\end{abstract}

%
\IEEEpeerreviewmaketitle

\section{Introduction}\glsresetall

\subsection{Motivation}

\Ac{mtl} provides state-of-the-art results in many applications of computer vision and \ac{nlp}~\cite{Ruder2017,Misra2016,Ruder2019}. 
In contrast to \ac{stl}, \ac{mtl} allows for leveraging knowledge between related tasks improving forecast results on the main task (in contrast to an auxiliary task) or all tasks. 
Simultaneously, learning multiple tasks increases the sample size and allows learning a more general representation~\cite{Ruder2017}, which in contrast to \ac{stl}, improves the forecast error.
Further, they typically reduce the computational effort.

Even though there are several articles evaluating the effectiveness of \ac{mtl} approaches for computer vision and \ac{nlp} problems, there is a limited number of comparative studies on applying \ac{mtl} architectures for regression and time series problems taking recent advances of \ac{mtl} into account.

\begin{figure}[t]
    \centering
    \includegraphics[width=\columnwidth]{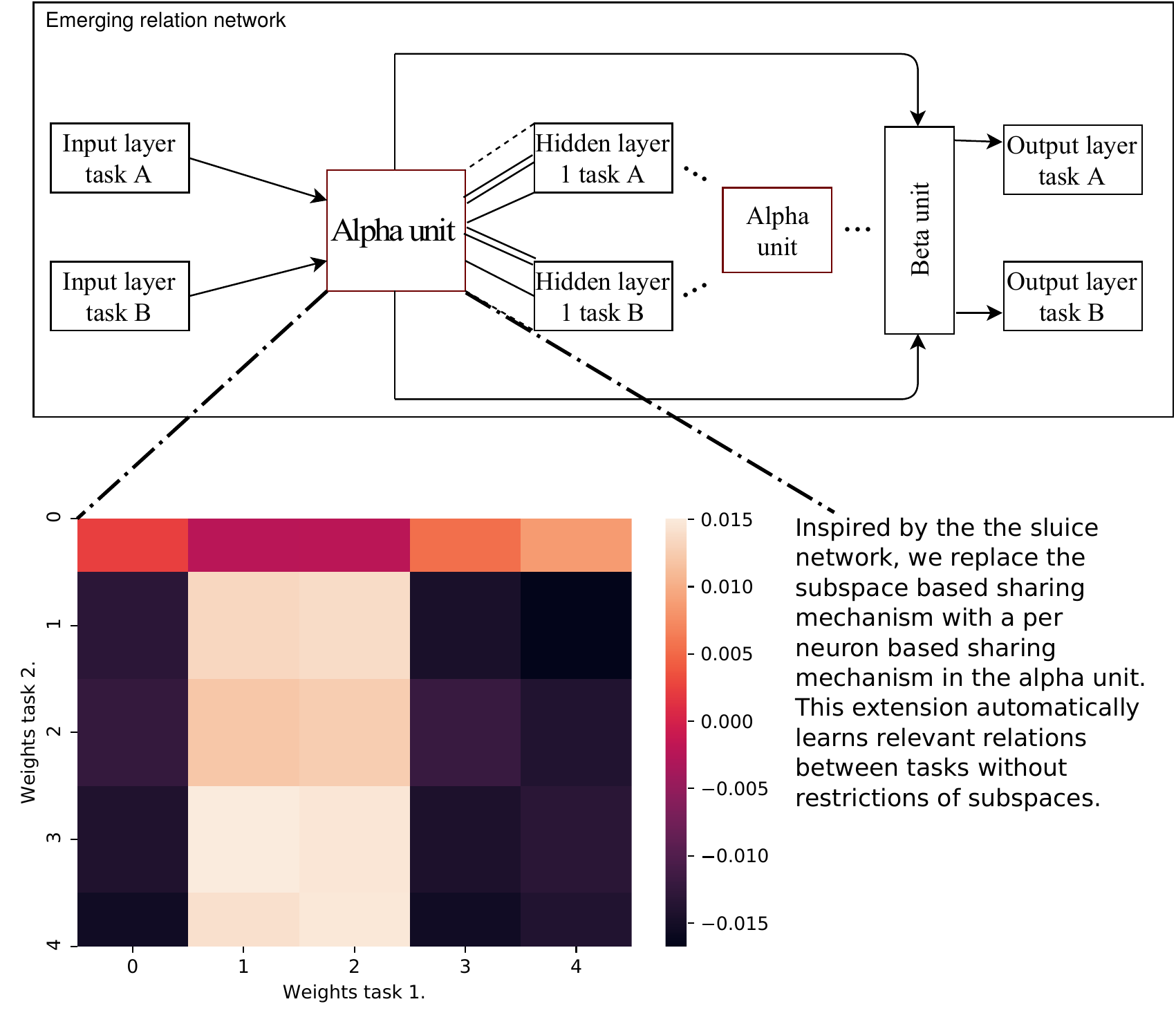}
    \caption{Schematic overview of the proposed emerging relation network. The network replaces the subspace based sharing mechanism of the sluice network with a per neuron based sharing mechanism in the alpha unit. Outputs of the alpha units are concatenated through skip layers, for each task and layer, as input to the beta unit, providing task specific forecasts.
    Different line types refer to (potential) different relations automatically emerged during training.
    Those different sharing mechanisms are present in the exemplary heatmap of weights.
    In the heatmap, it is noticeable that one group has negative weights; another group has positive weights, while others are around zero.}
    \label{fig_ern_scheme}
\end{figure}

One challenging prediction problem is the forecast of the expected power generation for renewable power plants. 
Typically, predicting power generation is a two step approach. 
The first step involves forecasting the weather features, such as wind speed or radiation, with a time step of up to $72$\si{\hour} in the future.
These forecasts from so-called \ac{nwp} are the input to the second step. 
In this step, weather features are mapped as a regression problem to the generated power of a solar or wind farm.
Overall, the process of forecasting the generated power, including the \ac{nwp}, is often considered a non-linear time series problem~\cite{Lange2006,Henze2020}. 

One problem is that the weather forecasts uncertainty increases with an increasing forecast horizon. 
Further, relations between different weather features, such as wind speed and air pressure, and between weather features and power generation, are non-linear.
Predicting the generated power is limited by the local weather information from the \ac{nwp}. 
However, an increased interest in renewable energy~\cite{fraunhofer2018windenergie} requires improved forecasts to maintain a stable power grid and for trading, while reducing the computational effort at the same time~\cite{Schwartz2019}.
As \ac{mtl} allows making use of information from other wind and solar parks, it allows reducing (local) uncertainty from weather predictions for a single park and decreases the forecast error, respectively. 
It also reduces the computational effort.

This article provides a comparative study of recent advances in \ac{mtl} for regression problems with an exemplary use-case in renewable energy.

\subsection{Main Contribution}

Therefore, this article provides a comparative study of the following recent and important \ac{mtl} architectures: 
\Ac{hps}, \ac{csn}, and \ac{sn} by comparing it to a \ac{mlp} model of similar size in an \ac{stl} setting\footnote{Source code is available under \url{https://git.ies.uni-kassel.de/mtl_regression/ern_and_te_for_mtl_regression_problems}.}.
Further, we introduce \textbf{\ac{ern}} that replace the subspace based sharing mechanism of \acp{sn} with a per neuron based sharing mechanism in the \textbf{alpha unit}.
This novel alpha unit allows the learning procedure to emerge relations automatically without a selection of the subspace hyperparameter. 
\FIG{fig_ern_scheme} gives the a schematic overview of the \ac{ern}. 
It also includes an example showing that relations and groups of relations emerge from the learning procedure, as highlighted in the heatmap of learned weights.

Additionally, we suggest a simple yet effective approach to model task specific information through an embedding layer in an \ac{mlp}, referred to as \textbf{task embedding}, see \FIG{fig_mlp_mtl}. 
Training and evaluating on a solar and a wind park dataset yields to the following significant results against the \ac{stl} \ac{mlp} baseline:

\begin{itemize}
    \item For the solar dataset, the task embedding achieves the best mean improvement, 
    \item the mean improvement of the \ac{ern} and the \ac{sn} are of similar magnitude.
    \item On the wind dataset, only the \ac{ern} achieves a significant improvement.
    \item Results suggest that the \ac{ern} is beneficial when tasks are only loosely related, and the prediction problem is more non-linear.
\end{itemize}

The remainder of this article is structured as follows. 
In \SEC{sec:related_work} we detail related work. 
\SEC{sec:method} outlines relevant deep learning architectures.
\SEC{sec:experimental_evaluation} describes the experimental design and evaluation results with respect to the \ac{stl} baseline.
Finally, we conclude our work and propose future work in \SEC{sec:conclusion}.
\section{Related Work}
\label{sec:related_work}

In the following section, we summarize \ac{mtl} in the field of deep learning, with a focus on computer vision and \ac{nlp}.
We limit the related work to approaches where the network learns the amount of joint and shared knowledge automatically.
Then in the next section, we detail related work for multi-task regression problems. 
We outline the limited utilization of deep learning methods in this area in general and its current focus on \ac{hps} \ac{mtl} architecture types.

\subsection{Deep Learning Based Multi-Task Learning}
In~\cite{Ruder2017}, an overview of novel methods in deep learning based \ac{mtl} is given, including their work on \ac{sn}.
The \ac{sn} is based on the \ac{csn} introduced in~\cite{Misra2016}.
The \ac{csn} learns a combination of task specific and universal representations through a linear combination for computer vision problems.
The \ac{sn} introduced in~\cite{Ruder2019} generalizes this idea by making use of skip layers and additional subspaces (with separate weights).
The additional subspaces provide a more fine-tuned separation between common and task specific sharing achieving excellent results in \ac{nlp}.

The formerly mentioned methods automatically learn what to share.
Approaches such as the deep relation networks~\cite{Long2017} require an at least partly pre-defined structure of the network.
In~\cite{Long2017}, several task specific layers follow some joint convolutional layers.
However, a prior to the separate layers allows them to automatically learn what to share, achieving good results for computer vision problems. 
Primarily, the network is considered an \ac{hps} architecture.
The authors of~\cite{Lu2017} use a greedy learning approach to dynamically create branches for task specific and joint layers for computer vision problems.
In~\cite{Sogaard2016} and~\cite{Hashimoto2017}, they focus on finding hierarchical network structures for \ac{nlp} problems.
The article \cite{Cipolla2018} introduces an approach for automatic weighting different tasks during training based on the uncertainty of a task. 
This approach adds additional complexity to the training process and is not related to a specific architecture type that we are interested in.

\subsection{Multi-Task Learning for Regression Problems}
As stated before, there is limited research on using deep learning architectures for multi-task regression problems.
In this section, we summarize articles in this area, for a detailed overview, refer to ~\cite{Ruder2017, Zhang2017}.
In several articles, the utilization of Gaussian processes models commonalities through an equal prior on the parameters of the Gaussian processes,~\cite{Ming2009, Rakitsch2013, MingChai2009, Zhang2009}. 
Other works make use of linear models modeling the relationship between related tasks, \cite{Lozano2012, Kim2010, Rai2012}. 
These models are computationally efficient but cannot model non-linear relationships between tasks.
In \cite{Tan2020}, the authors aim at modeling complex relations through shared weights of a least-squared support-vector regression.
In particular, the authors use this approach to predict heat, cooling, and gas loads.

The above articles disregard the capabilities of neural networks to learn shared knowledge automatically during training.
In contrast, the authors of \cite{Huang2020} combine knowledge from different network types through a fully connected neural network. 
In their goal to predict train delays, they have various kinds of data types. 
Therefore, the different data types are handled initially by either a long-term-short memory, an~\ac{mlp}, or a convolutional network. 
An \ac{mlp} combines the extracted knowledge from those networks to forecast the train delays.

The article~\cite{Liu2020} proposes an \ac{mtl} long short term memory network to forecast gas detection and concentration estimation of an intelligent electric sensing device simultaneously. 
Finally,~\cite{Dorado2020} aims at predicting wind power ramps. 
To make the best use of sparse data from different wind parks,~\cite{Liu2020} combines the knowledge through an \ac{hps} network. 
Further, an adapted Adam optimizer takes care of imbalanced data.

The article \cite{Vogt2019} gives good insights and excellent results in the utilization of \ac{mtl} architectures in a transfer learning setting with \ac{hps} networks. 
The authors also include a Bayesian variant of the proposed task embedding architecture.
However, the Bayesian variant adds additional complexity compared to our proposed approach and results are solely evaluated in the transfer learning setting.

The literature review shows that most of the work for multi-task regression models are either focusing on models without neural networks or are utilizing \ac{hps} networks. 
However, articles from the domain of computer vision and \ac{nlp} show noticeable results utilizing different types of \ac{hps} and \ac{sps} architectures.
\section{Methods}
\label{sec:method}
Typically in \ac{mtl}, we differentiate between \ac{hps} and \ac{sps} architectures.
Therefore, in \SEC{sec_hps} we give details on \ac{hps}, and in \SEC{sec_mlp} we propose a simple task encoding approach for \acp{mlp} in \ac{mtl} problems.
While this approach is considered an \ac{hps} architecture, \SEC{sec_csn} and \ref{sec_sn} introduce two recent advances from the fields of \ac{sps}.
In \SEC{sec_sn}, we introduce our novel approach \ac{ern}.

\subsection{Hard Parameter Sharing}\label{sec_hps}
\Ac{hps} networks are architectures, where several hidden layers learn a shared representation for all tasks. 
Additional task specific layers allow transferring knowledge from this representation to the task specific problem.

\subsection{MLP with Task ID Embedding}\label{sec_mlp}

\begin{figure}[t]
    \centering
    \includegraphics[width=0.95\columnwidth]{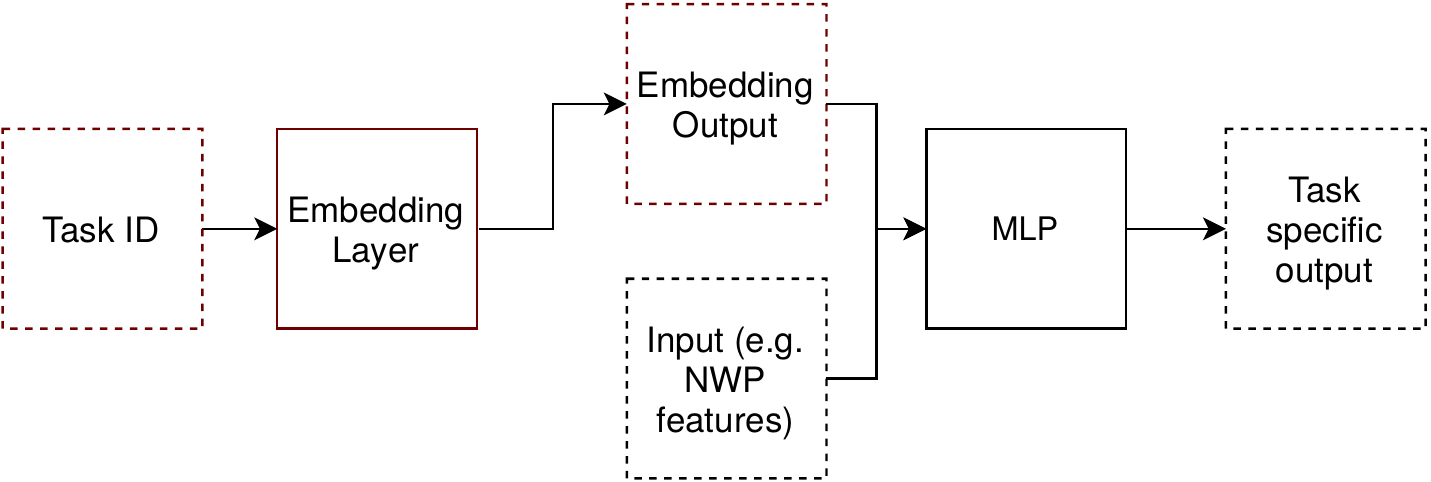}
    \caption{Task embedding for \ac{mlp} to create task specific predictions based on an hard parameter sharing architecture without separate layers. By encoding a task ID, for each task, through an embedding layer, the mlp learns task specific forecasts, while utilizing the data from all tasks to improve forecasts.}
    \label{fig_mlp_mtl}
\end{figure}

\FIG{fig_mlp_mtl} depicts our own approach for \ac{mtl} utilizing an \ac{mlp} and an embedding layer. 
Primarily, this approach is inspired by so-called word embeddings from the \ac{nlp} domain.
In the \ac{nlp} domain, word embeddings provide a continuous vector representation from a bag of words. 
The encoding through the embedding layer is learned, e.g., during a supervised training for sentiment classification.
In the end, after training, similar words, such as queen and king, have a similar representation, while not related words are far away from each other in the encoded representation. In the context of renewable energy, the similarity is, e.g., given when two wind parks have a similar mapping between wind speed and the generated power.

In the \ac{mtl} setting, we create a task ID for each task, additionally to other features. 
This task ID is the input to the embedding layer to encode task specific information.
The encoded task information is concatenated with other features from the \ac{nwp} as input to the \ac{mlp}.
The input layer for other features stays the same for all tasks.
During the supervised training, to forecast the generated power, the network learns (through backpropagation) the encoding while making use of other features to create task specific forecasts.

As all tasks share the same layers, except the task ID encoding, this approach can be considered as an \ac{hps} network.
In contrast to other \ac{hps} architectures, the task embedding architecture can reduce the number of parameters as it avoids additional separate layers.
Naturally, through the combined training of all tasks, the architecture makes use of data augmentation. 
Ideally, the learned task encoding, allows the network to make use of samples from a task \textit{A} for unknown weather situation in task \textit{B}.

\subsection{Cross-Stitch Network}\label{sec_csn}

In contrast to the previous two architectures, the \ac{csn}~\cite{Misra2016} is based on \ac{sps}.
In \ac{sps}, each task has a separate network learning a separate representation for each task.
In architectures such as \ac{csn} and \ac{sn}, information between the different networks is shared by so-called \textbf{alpha units}, as also visualized in our own approach in \FIG{fig_ern_scheme}.
Alpha units allow sharing information through a linear combination from one task to another learned during training through backpropagation. 

Equation~(\ref{eq_cs_alpha}) gives an \textit{example for two tasks} A and B at layer  $l \in {1,\ldots, L-1}$:

\begin{equation}\label{eq_cs_alpha}
    \begin{bmatrix}\tilde{h}^{\mathrm{T}}_{A,l} \\ \tilde{h}^{\mathrm{T}}_{B,l}  \end{bmatrix} = \begin{bmatrix} \alpha_{AA} & \alpha_{AB}\\ \alpha_{BA} & \alpha_{BB}  \end{bmatrix} \begin{bmatrix}h^{\mathrm{T}}_{A,l} & h^{\mathrm{T}}_{B,l}  \end{bmatrix},
\end{equation}

where the $\mathbf{\alpha}$ matrix is of size $\mathbb{R}^{2\times2}$ for those tasks.
The outputs $h_{A,l}$ and $h_{B,l}$ for the layer $l$ are concatenated and multiplied by the respective alpha unit at this layer.
The resulting linear combinations $\tilde{h}_{A,l}$ and $\tilde{h}_{B,l}$ are then utilized as input for the next layer of each task, similarly to \FIG{fig_ern_scheme}.
In this example for task \textit{A} and \textit{B}, the final output is then given by $\tilde{h}^{\mathrm{T}}_{A,L}$ and $\tilde{h}^{\mathrm{T}}_{B,L}$.

\subsection{Sluice Network}\label{sec_sn}
The authors of \cite{Ruder2019} extend the idea of \ac{csn} by two principles. 
First, they allow subspaces in layers (with different weights) to share knowledge between subspaces additionally besides tasks.
Naturally, this extends the \textbf{alpha units} by the number of subspaces, e.g., for two tasks (\textit{A} and \textit{B}) and two subspaces (1 and 2) the $\alpha$ matrix is  $\mathbb{R}^{4\times4}$ as given in the following equation:

\begin{equation*}\label{eq_cs_alpha}
    \begin{bmatrix}
        \tilde{h}^{\mathrm{T}}_{A_1,l} \\ \ldots \\ \tilde{h}^{\mathrm{T}}_{B_2,l}  \end{bmatrix} = 
        \begin{bmatrix} 
                \alpha_{A_1A_1} & \ldots & \alpha_{B_2A_1} \\ 
                \vdots & \ddots & \vdots \\ 
                \alpha_{A_1B_2} & \ldots &\alpha_{B_2B_2}  
    \end{bmatrix} 
\begin{bmatrix}h^{\mathrm{T}}_{A_1,l} & \ldots & h^{\mathrm{T}}_{B_2,l}  \end{bmatrix},
\end{equation*}
where $\alpha_{A_1B_2}$ refers to the alpha value of supspace one of task \textit{A} and supspace two of task B.
The output $\begin{bmatrix} \tilde{h}^{\mathrm{T}}_{A_1,l}, \ldots, \tilde{h}^{\mathrm{T}}_{B_2,l}  \end{bmatrix}^{\mathrm{T}}$ is the input to the next layer.

Further, the sluice network makes use of skip layers for the final prediction.
Therefore, the results $\tilde{h}_{t,l}$ of each task $t$ and layer $l$ are concatenated in a matrix. 
Through a linear combination of this matrix with the so-called \textbf{beta unit}, the calculation results in the final task specific output, similar to the beta unit of the \ac{ern} shown in \FIG{fig_ern_scheme}.
Again, the supervised training learns the values of the beta-unit automatically.

\subsection{Emerging Relation Network}\label{sec_ern}
Inspired by the work of~\cite{Misra2016,Ruder2019}, we replace the subspace based sharing mechanism of \acp{sn} with a per neuron based sharing mechanism in the \textbf{alpha unit}. 
Therefore, a matrix, where each dimension is equal to the summed number of neurons across all tasks at a layer $i$ replaces the alpha unit. 
This results in the following exemplarily calculation for two tasks with $m$ and $n$ neurons at layer $l$, respectively:

\begin{equation}\label{eq_ern_alpha}
\begin{bmatrix}\tilde{h}^{\mathrm{T}}_{1,l} \\ \tilde{h}^{\mathrm{T}}_{2,l}  \end{bmatrix} = 
    \begin{bmatrix} 
            \alpha_{1,1} & $\ldots$ & \alpha_{1,n}\\ 
            \ldots & $\ldots$ & \ldots  \\
            \alpha_{m, 1} & $\ldots$ & \alpha_{m,n}  
    \end{bmatrix} 
    \begin{bmatrix} h^{\mathrm{T}}_{1,l} & h^{\mathrm{T}}_{2,l}  
    \end{bmatrix},
\end{equation}
where $\alpha_{m,n}$ refers to neuron $m$ of the first task and neuron $n$ to the second task, 
$h_{1,l}$ is of size $m$, and $h_{2,l}$ of size $n$, respectively.

In our intuition, this provides the learning procedure to learn relevant relations between tasks without restrictions of the subspaces.
Additionally, this mechanism avoids selecting the hyperparameter for the number of subspaces.
Ideally, during the learning procedure, neurons or groups of neurons emerge that benefit from one task to one or more tasks. 
In the schematic overview in \FIG{fig_ern_scheme}, the exemplary weights show such a learned relation through the heatmap. 
In this heatmap, one group has negative weights;  another group has positive weights, while others are around zero, hence, sharing no information.

Note that this extension increases the number of parameters for the network.
The additional parameters increase with an increasing number of tasks, layers, and neurons in each layer.
However, it avoids selecting the additional subspace hyperparameter present in the \ac{sn}.
\section{Experimental Evaluation}
\label{sec:experimental_evaluation}

This section evaluates the experiments:

\begin{itemize}
    \item to quantify the improvement of deep learning based \ac{mtl} architectures compared to \ac{stl} models,
    \item to evaluate the task embedding for \ac{mlp}, and
    \item to evaluate the \ac{ern} against the \ac{sn}.
\end{itemize}

Therefore, we first give details on the design of the experiments, followed by the evaluation. Finally, we discuss the key results.

\subsection{Design of Experiments}

In our experiments, we compare a \ac{stl} model for each park against different \ac{mtl} approaches.
In particular, we evaluate the task embedding (MLPWP), an MLP without the task embedding (MLPNP), an \ac{hps} architecture, a \ac{csn}, and an \ac{sn}.
Finally, we evaluate our introduced \ac{ern}.
All networks include temporal information (e.g., hour of the day) via an embedding layer to incorporate information relevant for the time series forecast.
As a baseline, we use the single-park \ac{mlp} that also utilizes an embedding layer to consider the temporal information.

\subsubsection{Data}\label{sec_data}

In our experiments, we used a solar and a wind park dataset.
The solar dataset consists of $23$ parks, while the wind dataset consists of $15$ parks.
In this scenario, a task corresponds to the prediction of the generated power of a single park. 
Ideally, information sharing within the \ac{mtl} setting allows improving the overall forecast error across all tasks.

By making use of two datasets with different kinds of features and target variables, we show the strengths and weaknesses of the evaluated algorithms.
In both datasets, the uncertainty of the \ac{nwp} makes it challenging to predict the generated power.
As weather forecasts are valid for a larger area (a so-called grid-size of $2.8$~\si{\kilo\metre}), a mismatch between the forecast position and the actual placement of a park causes uncertainty.
Further, the uncertainty increases with an increasing forecast horizon of the weather prediction~\cite{Jens2019}.

The solar dataset has data from the beginning of $2015$ till the end of April $2016$.
The wind dataset covers the years $2015$ and $2016$.
In both cases, the year $2016$ is used as the test dataset. 
$80\%$ of the shuffled data from the year $2015$ is used as a training dataset and $20\%$ is used to find hyperparameters.

Each park in both datasets is standardized based on the training dataset.
By merging data from parks by their timestamps, we assure that information between tasks relate and we neglect non-overlapping timestamps.
As weather forecasts are only available in an hourly resolution, while the generated power is available in a $15$ minute resolution, the input features are linearly interpolated, resulting in four times more data samples per park.
All in all, this preprocessing results in $24328$ samples per park for training and validation for the wind dataset and $23040$ samples for testing.
For the solar dataset, $24796$ samples are used for training and validation and $8396$ samples for testing.

The solar dataset is generally considered a more straightforward problem compared to the wind park dataset. 
The relation between the input feature radiation and the generated power is mostly linear.
However, weather features influence each other non-linearly, making it still a challenging problem.
The solar dataset contains the following features: Temperature (at height $2$~\si{\metre}), geopotential, total surface cloud coverage, albedo surface, total surface precipitation, snow surface density, snow depth surface water equivalent, snow surface depth, mean diffuse and direct short wave surface radiation, direct radiation, diffuse radiation, radiation aggregated.
To improve the forecast quality, we add so-called time-shifted features from one hour in the past as well as one from the future from influencing weather features~\cite{Jens2018}.
For the solar dataset, we included in those sliding windows the following features: Direct and diffuse short wave radiation.

The wind dataset contains the following features: wind speed and wind direction (at height $32$~\si{\metre}, $73$~\si{\metre}, $122$~\si{\metre}), air pressure, temperature (at height $36$~\si{\metre}, $122$~\si{\metre}), relative humidity, unreduced ground pressure, pressure reduced, geopotential, and total precipitation.
Compared to the solar dataset, this dataset is more challenging as the relation between wind and the generated power is highly non-linear.
For the wind dataset, we included time-shifted features for all wind speeds and wind directions at different heights.

Further, we extracted the following temporal information (for both datasets) from the timestamps for all tasks: Hour of the day, week of the year, and day of the month.
As stated previously, encodes those features as input.

\subsubsection{Hyperparameters}

To ensure comparable network sizes, we follow the same principle for all networks. 
The first hidden layer size is equal to the number of input features multiplied by $10$.
Note that initially transforming the input features in a higher-dimensional space typically improves the performance~\cite{Sain2006}.
Afterward, in each layer, the number of neurons is reduced by $50\%$ to a minimum of $5$ neurons before the output.

For \ac{hps} networks, we include two task specific layers with sizes $5$ and $1$.
Utilizing Xavier as initialization, as a state-of-the-art method to initialize weights, we minimize the risk of exploding gradients~\cite{Glorot}.
We initialize alpha units with $0.9$ on the diagonal and $0.1/n$, where $n$ refers to the number of elements without the diagonal.
This balanced initialization assures that initially, tasks (\ac{csn}), subspaces (\ac{sn}), or neurons (\ac{ern}) have a large weight with themselves. 
Information between tasks, etc., have an initially smaller weight, hence sharing less information.

After each layer, except the output layer, we include (in the following order) a leaky rectified linear unit (with a slope of $0.01$), a batch normalization layer, and $50\%$ dropout.
For all embedding layers, we include a dropout of $25\%$.

For \ac{mtl} networks and the \ac{mlp} for each park, we use a batch size of $512$.
In the case we train a \ac{mlp} to forecast all parks, we use a batch size of $2048$ to reduce computational effort as the number of samples increases.
To accelerate the training, we train for $20$ epochs with a one-cycle learning rate scheduler~\cite{Smith2016} with a maximum learning rate of $0.01$ and cosine annealing.
To finetune the weights, we conclude the training with $100$ epochs and a small learning rate of \si{1e^{-4}}.
In all epochs, we shuffle the training data.
As training library we used \textit{pytorch}~\cite{pytorch} and \textit{fastai}~\cite{fastai}.

\subsubsection{Significance Test and Skill Score}

In the final results, we compare the forecast errors of each model and each park to baseline based on the \ac{rmse} on the test data.
On the solar dataset, we utilize the \textit{Wilcoxon signed-rank test} to show the significance of our models against the baseline.
As the wind dataset has only $15$ tasks and the Wilcoxon signed-rank test requires a sample size larger than $20$~\cite{scikit-learn}, we rely on a t-test.
Therefore, we test the differences between a model and the baseline for normality with the \textit{Shapiro-Wilk} test beforehand.
In all hypothesis tests, we use a  confidence level of $\alpha=0.01$.

To get insights into the amount of improvement, we use a mean skill score across all parks given by the following equation:
\begin{equation}
        \text{SkillScore} = \frac{1}{k} \sum_{k=1}^{k=K}{1-\frac{RMSE_{ref_k}}{RMSE_{baseline_k}}},
\end{equation}
where $k \in 1,\ldots,K$ refers to the current park, $RMSE_{ref_k}$ and $RMSE_{baseline_k}$ are the RMSE of the reference model and the baseline.
\subsection{Experimental Results}

\begin{figure}[b]
    \centering
    \subfloat[Pearson correlation coefficient between power generation of wind parks.]{\includegraphics[width=0.47\columnwidth]{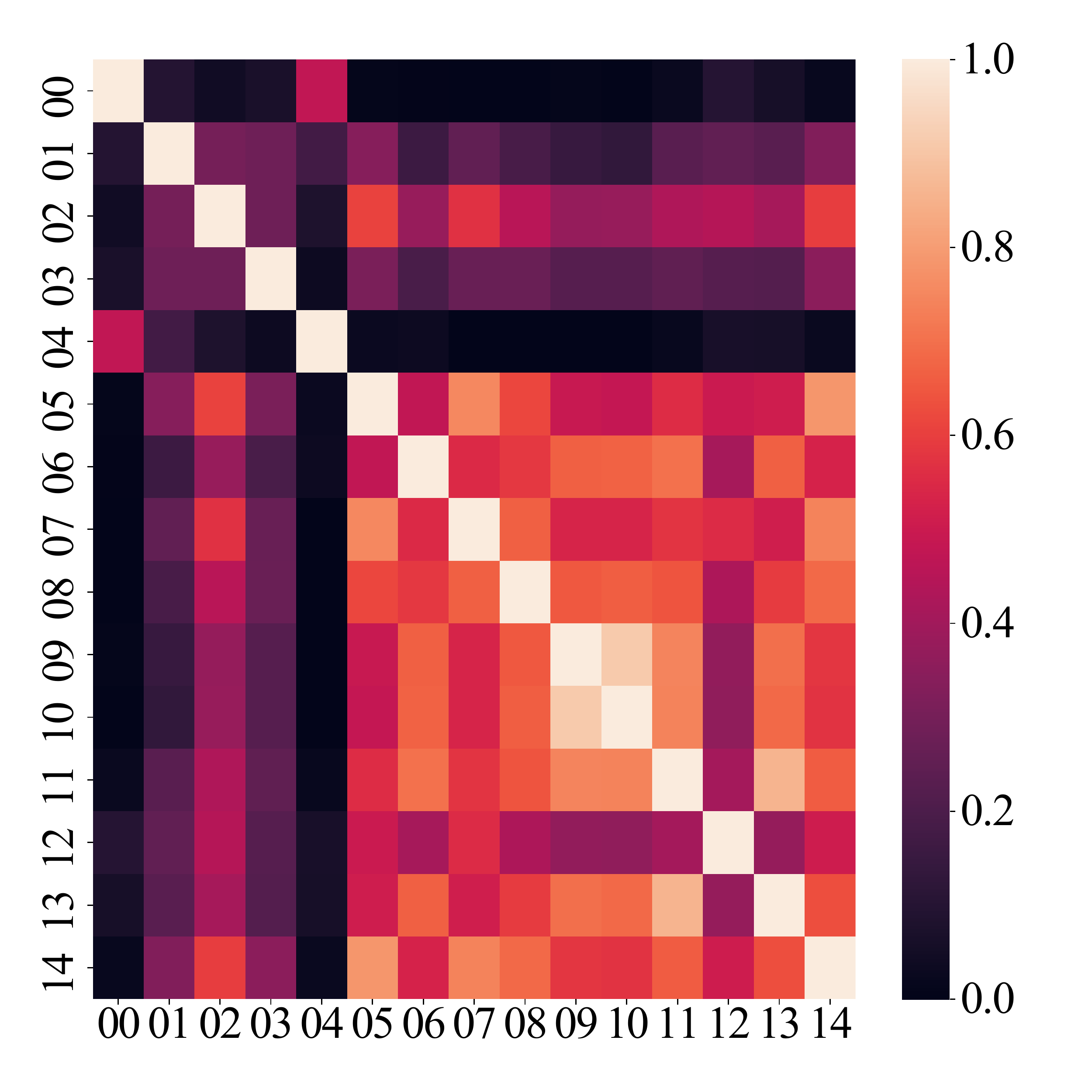}%
    \label{fig_pearson_wind}}
    \hfil
    \hfil
    \subfloat[Pearson correlation coefficient between power generation of solar parks.]{\includegraphics[width=0.47\columnwidth]{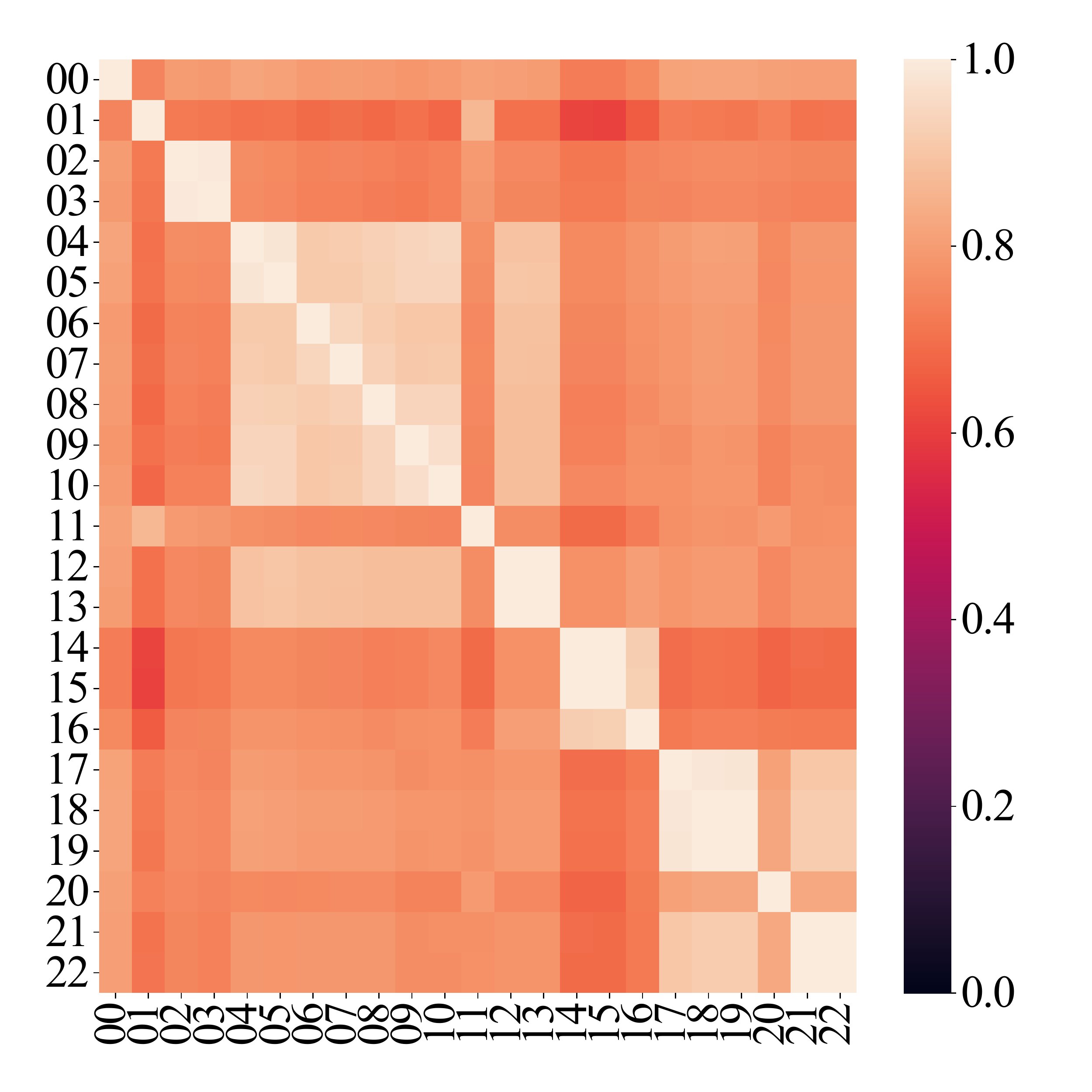}%
    \label{fig_pearson_solaw}}
    \caption{Pearson correlation coefficient of power generation for wind and solar parks calculated based on training and validation data.}
    \label{fig_pearson}
\end{figure}

This section details the evaluation results on the wind and the solar dataset described before.
As pointed out in the previous section, we train and evaluate several multi-task models on both datasets and compare it to the baseline. Note that the asterisk symbol marks significantly different models compared to the baseline. 

To get an initial impression on how different tasks relate to each other, we calculate the Pearson correlation on the training and validation data for the target variable between the different tasks. 
\FIG{fig_pearson} summarizes those results.
For the wind dataset, the Pearson correlation is between $0$ and $1$.
In \FIG{fig_pearson_wind}, at least six parks have a correlation below $0.4$.
In the case of solar parks in~\FIG{fig_pearson_solaw}, the correlation for all parks is between $0.6$ and $1$, indicating a higher correlation in contrast to the wind parks.

\begin{figure}[t]
    \centering
    \includegraphics[width=0.95\columnwidth]{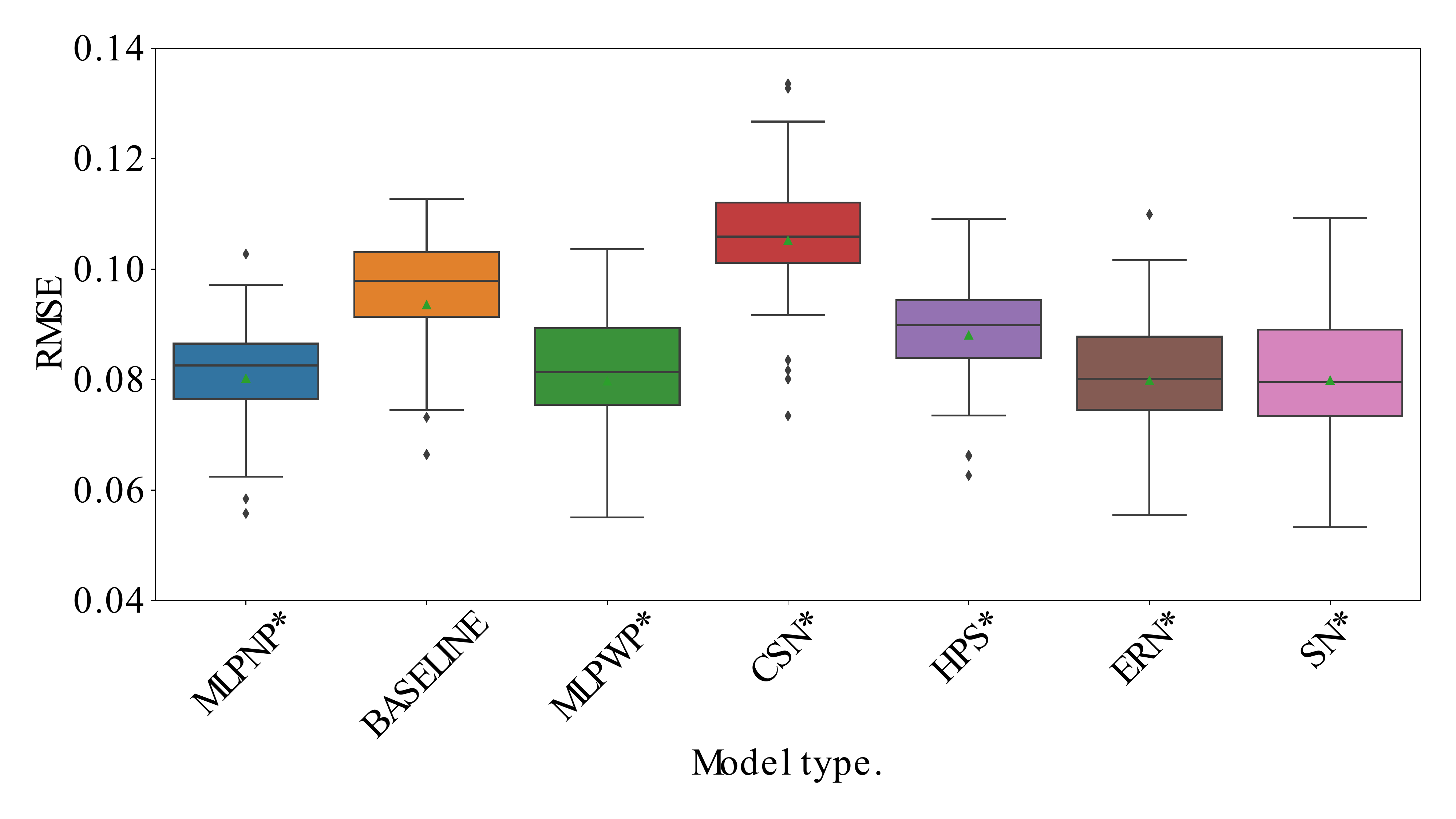}
    \caption{Evaluation results for the solar dataset. The asterisk symbol marks significantly different models compared to the baseline. Triangles indicate the mean. The boxplots include the following models: An MLP without task embedding (MLPNP), the \ac{stl} baseline, an MLP with task embedding (MLPWP), the cross-stitch network (CSN), the hard parameter sharing network (HPS), the emerging relation network (ERN), and the sluice network (SN).}
    \label{fig_pv_boxplot}
\end{figure}

\FIG{fig_pv_boxplot} summarizes the evaluation results for the solar dataset.
The task embedding improves the forecast error compared to the baseline. 
The \ac{csn} has a significantly worse error compared to the baseline.
The \ac{sn} model improves the quality of the forecast significantly and has, among all models, one of the best performances.
The proposed \ac{ern} also gives substantially better results than the \ac{stl} baseline. 
All models, except MLPWP and the SN, have outliers

\begin{table*}[!ht]
    \caption{RMSE results for each park and model for the solar dataset. The asterisk symbol marks significantly different models compared to the baseline. The table includes the following models: An MLP without task embedding (MLPNP), the \ac{stl} baseline, an MLP with task embedding (MLPWP), the cross-stitch network (CSN), the hard parameter sharing network (HPS), the emerging relation network (ERN), and the sluice network (SN).}\label{tbl_res_solar}
    \centering
    \footnotesize
    \begin{tabular}{lrrrrrrr}
        \toprule
        {ParkName} &  {BASELINE} &    {C{SN*}} &    {ERN*} &    {HPS*} &  {MLPNP*} &  {MLPWP*} &     {SN*} \\
        \midrule
        ParkPV00   &    0.0912 &  0.1051 &$\mathbf{  0.0738 }$&  0.0829 &  0.0767 &  0.0799 &  0.0781 \\
        ParkPV01   &    0.0732 &  0.0835 &  0.0572 &  0.0735 &  0.0584 &  0.0552 &$\mathbf{  0.0533 }$\\
        ParkPV02   &    0.0982 &  0.1023 &$\mathbf{  0.0801 }$&  0.0888 &  0.0842 &  0.0852 &  0.0834 \\
        ParkPV03   &    0.0950 &  0.1327 &  0.0774 &  0.0883 &  0.0771 &  0.0775 &$\mathbf{  0.0749 }$\\
        ParkPV04   &    0.0745 &  0.0817 &$\mathbf{  0.0587 }$&  0.0661 &  0.0627 &  0.0631 &  0.0609 \\
        ParkPV05   &    0.1039 &  0.1090 &$\mathbf{  0.0838 }$&  0.0942 &  0.0854 &  0.0897 &  0.0894 \\
        ParkPV06   &    0.0988 &  0.1115 &  0.0916 &  0.0929 &$\mathbf{  0.0876 }$&  0.0877 &  0.0933 \\
        ParkPV07   &    0.0664 &  0.0734 &  0.0554 &  0.0626 &  0.0557 &  0.0550 &$\mathbf{  0.0544 }$\\
        ParkPV08   &    0.0937 &  0.0999 &$\mathbf{  0.0763 }$&  0.0849 &  0.0798 &  0.0807 &  0.0796 \\
        ParkPV09   &    0.0664 &  0.0916 &  0.0639 &  0.0663 &  0.0732 &$\mathbf{  0.0607 }$&  0.0633 \\
        ParkPV10   &    0.1009 &  0.1059 &  0.0813 &  0.0926 &  0.0825 &  0.0820 &$\mathbf{  0.0788 }$\\
        ParkPV11   &    0.1028 &  0.1067 &$\mathbf{  0.0822 }$&  0.0946 &  0.0841 &  0.0895 &  0.0847 \\
        ParkPV12   &    0.0979 &  0.1139 &$\mathbf{  0.0827 }$&  0.0909 &  0.0835 &  0.0866 &  0.0839 \\
        ParkPV13   &    0.0962 &  0.1028 &  0.0752 &  0.0898 &  0.0783 &  0.0766 &$\mathbf{  0.0745 }$\\
        ParkPV14   &    0.1058 &  0.1186 &  0.0916 &  0.1070 &  0.0972 &  0.0891 &$\mathbf{  0.0886 }$\\
        ParkPV15   &    0.1127 &  0.1335 &  0.1099 &  0.1091 &$\mathbf{  0.1027 }$&  0.1036 &  0.1092 \\
        ParkPV16   &    0.0943 &  0.1126 &  0.0756 &  0.0888 &  0.0762 &  0.0742 &$\mathbf{  0.0722 }$\\
        ParkPV17   &    0.1039 &  0.1057 &  0.1016 &  0.0999 &$\mathbf{  0.0951 }$&  0.0956 &  0.1019 \\
        ParkPV18   &    0.0915 &  0.1031 &  0.1012 &  0.0894 &$\mathbf{  0.0783 }$&  0.0786 &  0.0982 \\
        ParkPV19   &    0.1013 &  0.1267 &$\mathbf{  0.0862 }$&  0.0935 &  0.0875 &  0.0903 &  0.0884 \\
        ParkPV20   &    0.1047 &  0.1105 &$\mathbf{  0.0893 }$&  0.0975 &  0.0906 &  0.0932 &  0.0905 \\
        ParkPV21   &    0.0746 &  0.0801 &  0.0608 &  0.0760 &  0.0624 &  0.0584 &$\mathbf{  0.0570 }$\\
        ParkPV22   &    0.1033 &  0.1079 &  0.0794 &  0.0949 &  0.0856 &  0.0813 &$\mathbf{  0.0780 }$\\
        \bottomrule
        \bottomrule
        \textbf{SkillScore} &    0.0000 & -0.1281 &  0.1477 &  0.0573 &  0.1410 &$\mathbf{  0.1496 }$&  0.1485 \\
        \bottomrule
        \end{tabular}
\end{table*}

More detailed results for each park are given in \TBL{tbl_res_solar}.
The table presents the \ac{rmse} for each park and model and highlights the best one in bold. 
Results are summarized through the mean skill score at the end of the table compared to the baseline.
Interestingly, the task embedding model yields the best results regarding the skill score.
In particular, the model has an improvement of $14.96\%$.
The MLPNP model, without the additional task embedding, has a lower improvement of $14.1\%$.
The \ac{hps} architecture has among the smallest improvement with $5.73\%$.
The SN model has an improvement of $14.85\%$.
Finally, the ERN has a skill score of $14.77\%$. 

\begin{figure}[!h]
    \centering
    \includegraphics[width=0.95\columnwidth]{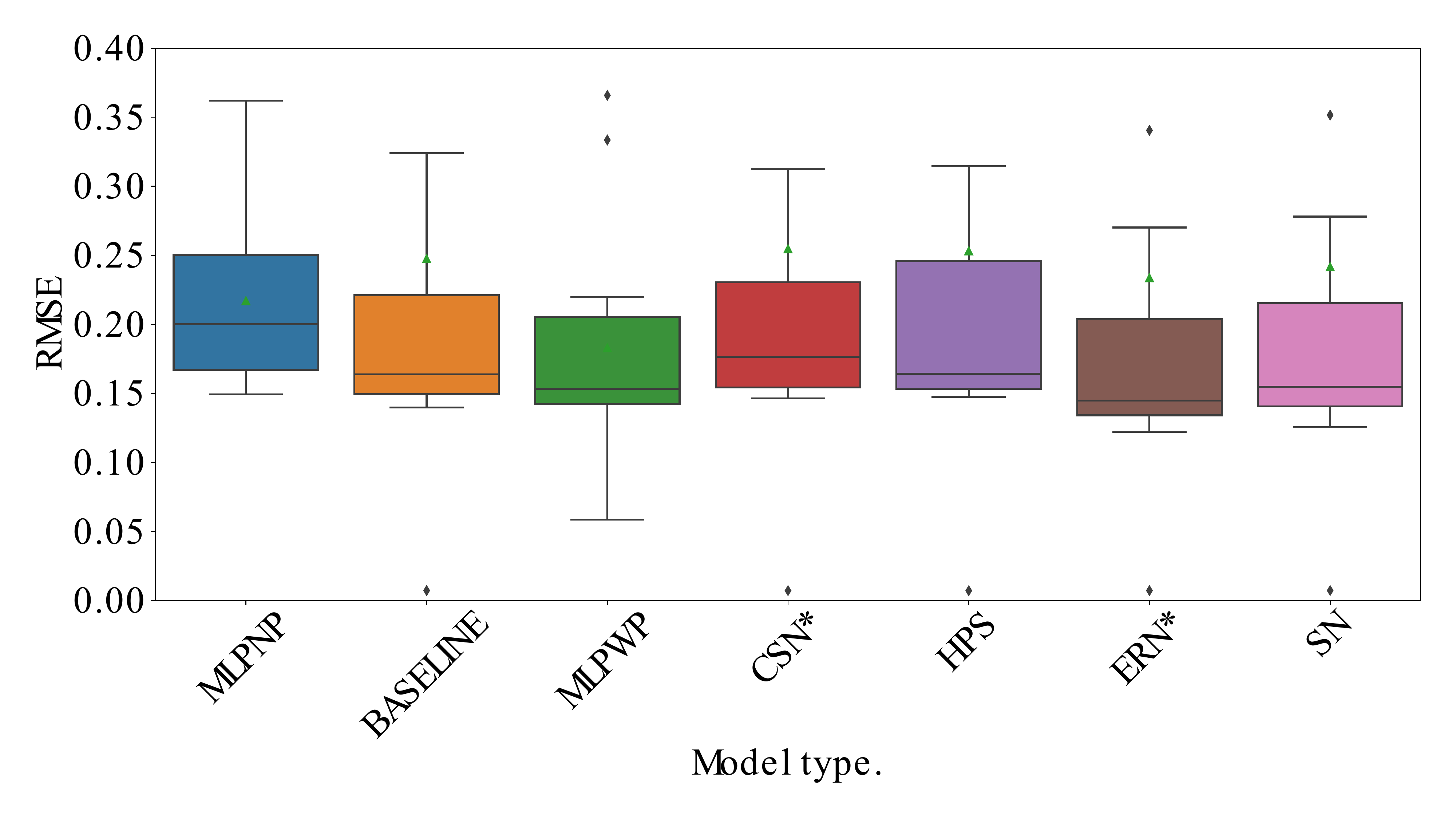}
    \caption{Evaluation results for the wind dataset. The asterisk symbol marks significantly different models compared to the baseline. Triangles indicate the mean. The boxplots include the following models: An MLP without task embedding (MLPNP), the \ac{stl} baseline, an MLP with task embedding (MLPWP), the cross-stitch network (CSN), the hard parameter sharing network (HPS), the emerging relation network (ERN), and the sluice network (SN).}
    \label{fig_wind_boxplot}
\end{figure}

\FIG{fig_wind_boxplot} summarizes results on the wind park dataset.
The CSN is worse than the baseline.
Our approach, the ERN model, outperforms the baseline and has two outliers.
All other models are not statistically different from the baseline.

\begin{table*}[!htb]
    \caption{RMSE results for each park and model for the wind dataset. The asterisk symbol marks significantly different models compared to the baseline. The table includes the following models: An MLP without task embedding (MLPNP), the \ac{stl} baseline, an MLP with task embedding (MLPWP), the cross-stitch network (CSN), the hard parameter sharing network (HPS), the emerging relation network (ERN), and the sluice network (SN).}\label{tbl_res_wind}
    \centering
    \footnotesize
    \begin{tabular}{lrrrrrrr}
        \toprule
        {ParkName} &  {BASELINE} &    {C{SN}*} &    {ERN*} &     {HPS} &   {MLPNP} &   {MLPWP} &      {SN} \\
        \midrule
        ParkWind00 &    0.1633 &  0.1749 &$\mathbf{  0.1330 }$&  0.1622 &  0.1710 &  0.1503 &  0.1476 \\
        ParkWind01 &$\mathbf{    0.1553 }$&  0.1558 &  0.1592 &  0.1578 &  0.2047 &  0.1559 &  0.1640 \\
        ParkWind02 &    0.2054 &  0.2103 &$\mathbf{  0.1737 }$&  0.1978 &  0.2235 &  0.1919 &  0.1943 \\
        ParkWind03 &    0.1637 &  0.1763 &$\mathbf{  0.1448 }$&  0.1641 &  0.1766 &  0.1531 &  0.1493 \\
        ParkWind04 &    0.2858 &  0.2956 &  0.2701 &  0.2879 &  0.2412 &$\mathbf{  0.2006 }$&  0.2780 \\
        ParkWind05 &    1.2202 &  1.2266 &  1.2083 &  1.2384 &$\mathbf{  0.3620 }$&  0.3658 &  1.2090 \\
        ParkWind06 &    0.1647 &  0.1772 &$\mathbf{  0.1376 }$&  0.1623 &  0.1730 &  0.1525 &  0.1548 \\
        ParkWind07 &    0.1397 &  0.1463 &$\mathbf{  0.1386 }$&  0.1473 &  0.1491 &  0.1421 &  0.1448 \\
        ParkWind08 &    0.1541 &  0.1732 &$\mathbf{  0.1351 }$&  0.1687 &  0.1629 &  0.1420 &  0.1360 \\
        ParkWind09 &    0.0072 &  0.0071 &  0.0071 &$\mathbf{  0.0070 }$&  0.2001 &  0.0586 &  0.0072 \\
        ParkWind10 &    0.3240 &$\mathbf{  0.3125 }$&  0.3404 &  0.3146 &  0.3314 &  0.3335 &  0.3515 \\
        ParkWind11 &    0.2271 &  0.2377 &$\mathbf{  0.1928 }$&  0.2214 &  0.2595 &  0.2101 &  0.2132 \\
        ParkWind12 &    0.1426 &  0.1502 &$\mathbf{  0.1220 }$&  0.1483 &  0.1510 &  0.1344 &  0.1255 \\
        ParkWind13 &    0.2150 &  0.2233 &$\mathbf{  0.2147 }$&  0.2702 &  0.2994 &  0.2196 &  0.2175 \\
        ParkWind14 &    0.1446 &  0.1525 &$\mathbf{  0.1271 }$&  0.1485 &  0.1497 &  0.1340 &  0.1312 \\
        \bottomrule
        \bottomrule
        \textbf{SkillScore} &    0.0000 & -0.0409 &$\mathbf{  0.0774 }$& -0.0256 & -1.8287 & -0.3790 &  0.0354 \\
        \bottomrule
        \end{tabular}
\end{table*}

\TBL{tbl_res_wind} details the results from the boxplot. 
For ten parks, the ERN achieves the best results resulting in an overall improvement of $7.74\%$.
The SN has a small improvement of $3.54\%$; however, this result is not significant.
All other models have a negative skill score and are not significantly better than the baseline.
Also note, that the RMSEs of most models for \textit{ParkWind05} are rather large and indicate an outlier.

\subsection{Discussion}

Generally, in our results, we can see that \ac{mtl} architectures improve upon a \ac{stl} model of similar model size.
Depending on the dataset and the model, the results vary. 
For solar dataset, the task embedding leads to the best improvement in terms of the skill score.
This result is partly surprising as it is considered an \ac{hps} network forcing the learning process to learn a common representation instead of a task specific one.
However, most solar parks have a high Pearson correlation, see \FIG{fig_pearson_solaw}.
This strong correlation is probably beneficial for a common representation across all tasks resulting in the best results.
The results of the MLPNP also support this observation. 
The mean skill score of above $14\%$ suggests that the data augmentation through combined training of all tasks already improves upon the \ac{stl} architecture for the solar dataset even without task specific information or layers.

Results of the \ac{sps} architectures, except the \ac{csn}, lead to a significant improvement compared to the baseline.
However, the mean skill score is slightly worse compared to the task embedding architecture.
The improvement of \ac{sn} and \ac{ern} are of similar magnitude.
Suggesting that it is beneficial to utilize the \ac{ern} to avoid selection of the subspace hyperparameter of the \ac{sn}.

In the wind dataset, only the \ac{sn} achieves a significant improvement upon the \ac{stl} baseline.
This effect is explainable by the loosely related tasks, see \FIG{fig_pearson_wind}.
During training, the neuron based sharing mechanism allows the \ac{ern} to learn a representation and a sharing mechanism that is beneficial for the small correlation between tasks automatically.
The \ac{sn} results could potentially improve with a different number of subspaces. 
However, this would require a hyperparameter search, which the sharing mechanism of the \ac{ern} avoids.

\section{Conclusion and Future Work}
\label{sec:conclusion}

In our article, we successfully showed the improvement of \ac{mtl} architectures upon \ac{stl} models.
Therefore, we quantified the improvement employing the mean skill score based on the RMSE for a solar and wind park dataset.
We also showed their significance against an MLP baseline for each park.
Further, we suggest two new architectures that help in tackling different challenges in \ac{mtl} predictions. 

The task embedding architecture provides a simple and effective method when tasks are strongly correlated and a common representation is beneficial. 
In contrast to \ac{sps} architectures, this \ac{hps} model limits the number of required parameters while achieving the best significant results on the solar dataset.
As complex models require extensive computational resources and are contrary to climate goals~\cite{Schwartz2019}, the task embedding provides are a robust model that improves the forecast error while minimizing the computational effort.

The proposed adaption of the \ac{sn} through a per neuron based sharing mechanism allows the \ac{ern} to achieve results of similar magnitude on the solar dataset compared to the task embedding and the \ac{sn}.
Results on the wind dataset show that the automatic learning process is superior compared to its predecessor, the \ac{sn}, as it is the only one improving upon the baseline on the wind dataset.

A future goal is to incorporate additional data for the solar dataset to increase the expressiveness of the results. Further, we aim at utilizing the learned knowledge from \ac{mtl} architectures for predictions of parks with limited historical data.

\section*{Acknowledgment}
\small
This work was supported within the project Prophesy (0324104A) funded by BMWi (Deusches Bundesministerium für Wirtschaft und Energie / German Federal Ministry for Economic Affairs and Energy).



%


\vspace{-0.5em}
\bibliographystyle{IEEEtran}
\bibliography{references.bib}

\end{document}